\documentclass[10pt,twocolumn,letterpaper]{article}

\usepackage[pagenumbers]{cvpr} % To force page numbers, e.g. for an arXiv version

\usepackage{graphicx}
\usepackage{amsmath}
\usepackage{amssymb}
\usepackage{booktabs}
\usepackage{multicol}
\usepackage{multirow}
\usepackage{wrapfig}
\usepackage{enumitem}
\usepackage{array}
\usepackage{gensymb}
\usepackage{siunitx}
\usepackage{setspace}
\usepackage{xspace}
\usepackage{pifont}

\newcolumntype{?}{!{\vrule width 1pt}}

\usepackage[pagebackref,breaklinks,colorlinks]{hyperref}

\usepackage[capitalize]{cleveref}
\crefname{section}{Sec.}{Secs.}
\Crefname{section}{Section}{Sections}
\Crefname{table}{Table}{Tables}
\crefname{table}{Tab.}{Tabs.}

\newcommand{\csr}{CSR\xspace}
\newcommand{\cmark}{\ding{51}}
\newcommand{\xmark}{\ding{55}}

\def\cvprPaperID{2464} % *** Enter the CVPR Paper ID here
\def\confName{CVPR}
\def\confYear{2022}

\newcommand{\mypara}{\noindent\textbf}
\begin{document}

\title{Continuous Scene Representations for Embodied AI}

\author{Samir Yitzhak Gadre$^1$\thanks{Work done while SYG was a research intern at AI2. Correspondence to \texttt{sy@cs.columbia.edu}}\hspace{10mm}
Kiana Ehsani$^2$\hspace{10mm}
Shuran Song$^1$\hspace{10mm}
Roozbeh Mottaghi$^{2,3}$\\
$^1$ Columbia University\hspace{5mm}
$^2$ Allen Institute for AI\hspace{5mm}
$^3$ University of Washington\\
\href{https://prior.allenai.org/projects/csr}{prior.allenai.org/projects/csr}
}
\maketitle

%%%%%%%%% ABSTRACT
\begin{abstract}

We propose Continuous Scene Representations (\csr), a scene representation constructed by an embodied agent navigating within a space, where objects and their relationships are modeled by continuous valued embeddings. Our method captures feature relationships between objects, composes them into a graph structure on-the-fly, and situates an embodied agent within the representation. Our key insight is to embed pair-wise relationships between objects in a latent space. This allows for a richer representation compared to discrete relations (e.g., \textsc{[support]}, \textsc{[next-to]}) commonly used for building scene representations. \csr can track objects as the agent moves in a scene, update the representation accordingly, and detect changes in room configurations. Using \csr, we outperform state-of-the-art approaches for the challenging downstream task of visual room rearrangement, without any task specific training. Moreover, we show the learned embeddings capture salient spatial details of the scene and show applicability to real world data.
A summery video and code is available at \href{https://prior.allenai.org/projects/csr}{prior.allenai.org/projects/csr}.

\end{abstract}
%%%%%%%%% ABSTRACT

%%%%%%%%% BODY TEXT

\section{Introduction}
To operate within a scene, embodied agents require
a comprehensive representation of their surroundings. Such
perceptual understanding should not be limited to determining object identities, but should rather capture relationships
between objects and between the agent and its surroundings.
An expressive scene representation should also allow for downstream task completion of interactive tasks without additional training (i.e., zero-shot inference).

\begin{figure}[tp]
    \centering
    \includegraphics[width=19pc]{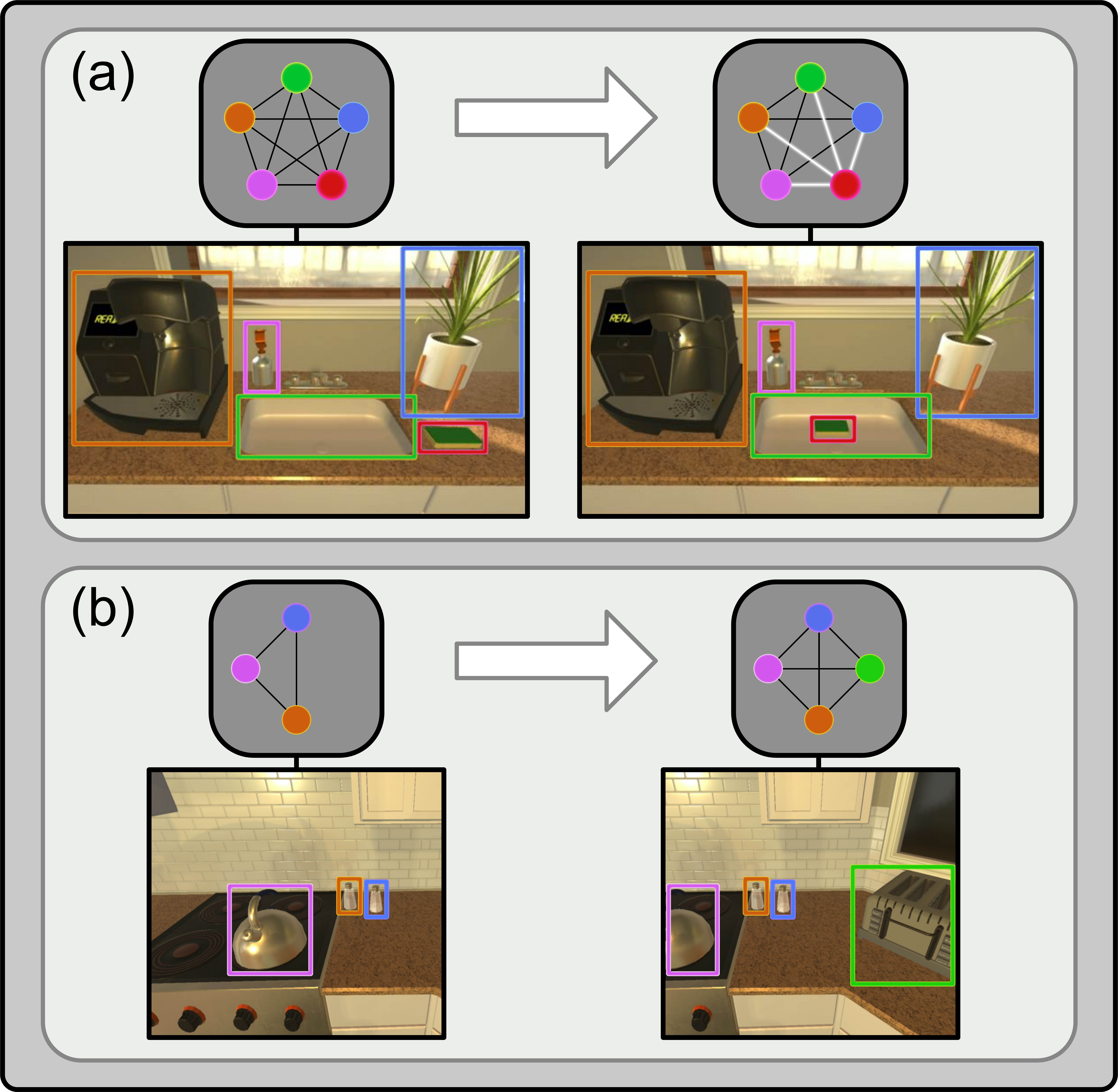}
    \vspace{-0.2cm}
    \caption{
        \textbf{Representing relationships.}
        Our work, Continuous Scene Representations, encodes object relationships into a graph with nodes and edges represented as feature vectors.
        \textbf{(a)} If an object is shuffled between two trajectories, different embeddings allow for change detection.
        \textbf{(b)} As the agent moves within a trajectory, new nodes and edges are populated.
        }
    \label{fig:teaser}
    \vspace{-3mm}
\end{figure}

Towards building such representations, scene graphs \cite{Johnson2015CVPR,zhao2013scene,armeni20193d} are a candidate as they provide a compact and explicit description of a scene.
In a typical scene graph pipeline, it is common to define a set of relationship labels, use them to manually annotate connections between objects in frames, and train a model to infer target graphs.
However, once labels are defined, there are inherent limitations.
Once trained, models are restricted to a fixed set of relationships.

Even if all relationships are modeled, how well can pre-defined discrete symbols represent the complex relationships between objects? Consider \textsc{[support]}, a common semantic relation used in the literature \cite{zellers2018neural,yang2018graph,li2017scene}. The relationship ``table supports mug" indicates the mug is on the table, but it does not capture \emph{where}. In practice, the agent behavior may depend on this information.

Beyond the inherent limitations of modeling relationships as discrete symbols, scene graphs used in the literature are typically static \cite{xu2017scene,ji2020action,lu2016visual} (i.e., they represent a snapshot of a scene or a bundle of frames).
In embodied settings, new objects are observed as an agent explores, and the scene representation should update on-the-fly.
If an agent returns to a position, it should determine if objects have moved.

Inspired by these observations, we develop a scene representation that is more suitable for embodied AI tasks.
We propose Continuous Scene Representations (\csr), a novel approach to construct a scene representation from egocentric RGB images as an embodied agent explores a scene.
To address limitations of traditional scene graphs, our goal is to represent relations between objects as continuous vectors and update the representation on-the-fly as the agent moves.
To enable downstream interactive task execution without additional training, we also present a simple strategy for planning with respect to the representation.

Constructing such a scene representation interactively poses various challenges.
The graph should accommodate the new objects, and relationships with other objects should be inferred.
A successful algorithm should also determine when detections correspond to the same object even in different views.
To tackle these challenges, our idea is to learn \emph{object relational embeddings} via a contrastive loss to represent the nodes and edges of a scene representation.
In this scheme, modeling is not constrained to a pre-defined set of symbols.
The agent maintains a memory of previously encountered embeddings.
As the agent extracts new embeddings from egocentric observations, it compares them to the memory to determine which embeddings are new and which already exist.
This can be used to detect changes in the scene and update the representation
as shown in Fig. \ref{fig:teaser}.

We perform experiments using the AI2-THOR \cite{ai2thor} framework
and the YCB-Video dataset \cite{xiang2018posecnn}.
We support experimentally that
(1) \emph{without any task specific training}, a simple planning approach that employs \csr as the underlying representation, outperforms a map-based, reinforcement and imitation learning baseline trained directly on the task of room rearrangement~\cite{batra2020rearrangement,weihs2021visual}.
(2) \csr is able to capture commonly used discrete relations that can be extracted via linear probes.
(3) \csr effectively captures spatial relationship between objects within a scene.
(4) Without any fine-tuning or hyperparameter tuning, a \csr trained on AI2-THOR is able to track objects over time in real world YCB-Video~\cite{xiang2018posecnn}, which contains objects unseen during training.
\section{Related Work}

\noindent\textbf{Scene graphs.} There are various methods in the literature that use scene graphs to develop richer scene understanding models.
It is common to build scene graphs for static images \cite{lu2016visual,li2017scene,xu2017scene,dai2017detecting,zellers2018neural,yang2018graph}.
One of the major issues of these approaches is the lack of temporal information (i.e., there is no mechanism to identify or track changes in a scene over time).
To overcome this issue, other approaches create scene graphs from videos \cite{shang2017video,tsai2019video,liu2020beyond,ji2020action,cong2021spatial,ost2021neural}.
These types of scene graphs capture temporal information; however, they are created using pre-recorded videos, hence, not suitable for embodied tasks that involve observations dependent on on-the-fly actions.
The mentioned image-based and video-based scene graphs capture only 2D relations between objects in a scene. 
However, others propose to create 3D scene graphs that encode 3D relationships between the components of a scene \cite{armeni20193d,wald2020learning,zhou2019scenegraphnet,chen2019holistic++,fisher2011characterizing}.
Our method is closer to the approaches that create scene graphs in embodied or physics based settings \cite{rosinol20203d,du2020learning,zhu2021hierarchical,li2021embodied,Battaglia2016InteractionNF}.
There are two important differences between our work and these approaches.
First, we do not have a pre-defined set of relationships (e.g., \textsc{[support]}, \textsc{[next-to]}).
Instead, we represent the relationships by a learned embedding. 
Second, our approach does not require object category information at inference. 

\noindent\textbf{Mapping and localization.}
Scene graphs are a form of abstraction for metric-semantic maps, so we discuss examples of mapping work.
Classical mapping and localization work in the robotics literature \cite{mur2017orb,engel2014lsd,davison2007monoslam} relies on low-level geometric features and does not encode semantic information.
There are object-based maps that encode object-level semantics and the relationships \cite{bao2012semantic,salas2013slam++,bowman2017probabilistic}.
However, they still operate based on the assumption of static environments.
Dynamic SLAM approaches \cite{runz2017co,xu2019mid,strecke2019fusion} have been proposed to handle moving objects in a scene.
However, they primarily focus on table-top scenes.
Recently, approaches have been designed to handle more complex dynamic scenes \cite{Rosinol21ijrr-Kimera,wong2021rigidfusion}.
There are few major differences with this work. (1) Our method produces an abstract sparse representation rather than a dense representation.(2) We create a continuous representation of the scene. 
(3) Methods such as \cite{wong2021rigidfusion} require a full sequence of observations as input, which is not suitable for embodied applications since the observations vary depending on the taken actions.  

\noindent\textbf{Embodied tasks.} Recently, various embodied tasks such as navigation \cite{yang2018visual,wortsman2019learning,Wijmans2020DDPPOLN,perez2021robot,chaplot2020learning,objectnav}, instruction following \cite{anderson2018vision,Shridhar2020ALFREDAB}, embodied question answering \cite{Das2018EmbodiedQA,Gordon2018IQAVQ}, object manipulation \cite{ehsani2021manipulathor,shen2020igibson,gan2021threedworld,gadre2021act}, and room rearrangement \cite{batra2020rearrangement,weihs2021visual} have been of interest.
These tasks benefit substantially from some form of memory to store the state of the scene.
Neural graph-based \cite{Savarese-RSS-19,savinov2018semi,wu2019bayesian,chaplot2020neural} and dense representations \cite{gupta2017cognitive,chaplot2020learning} have been proposed to construct the map of the environment.
These methods mostly focus on creating occupancy maps of free space and obstacles or the reachability of different nodes.
In our approach, we propose a scene representation that encodes object relations by feature embedding and evolves based on new observations and agent actions. 

\section{Scene Representation}
\label{sec:approach}
\begin{figure}[tp]
    \centering
    \includegraphics[width=\linewidth]{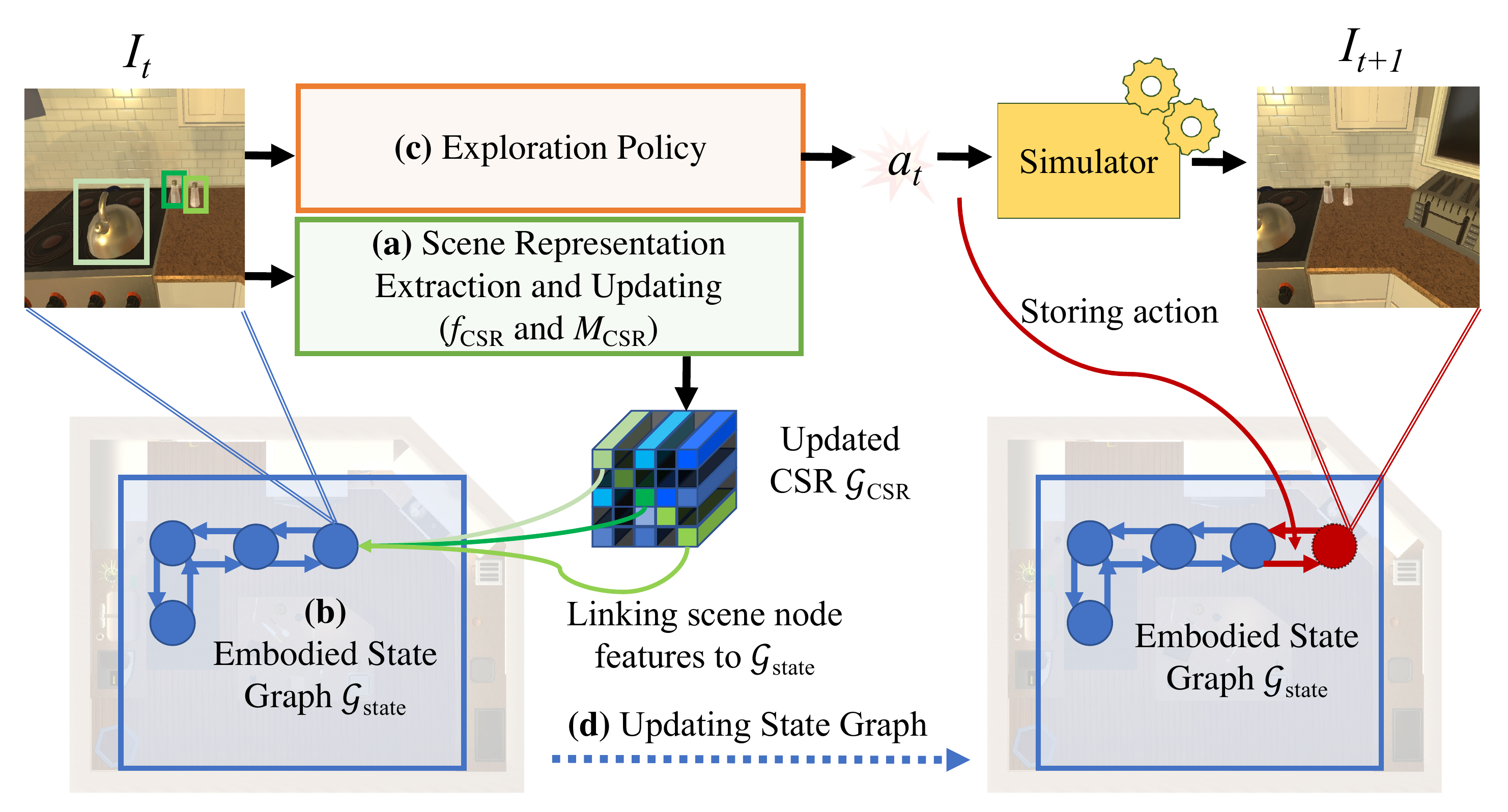}
    \caption{
        \textbf{Overview.}
        \textbf{(a)} Given an egocentric observation $I_t$, the scene representation module extracts nodes and edges to update the continuous scene representation (\csr).
        \textbf{(b)} References to observed nodes in the current image are stored in an embodied state graph, which links agent poses to \csr nodes that are seen from that pose.
        \textbf{(c)} An exploration policy infers an action to more comprehensively observe the scene and build a more complete \csr.
        \textbf{(d)} As the agent moves, the embodied state graph is updated with the action that leads to the next agent pose.
    }
    \label{fig:overview}
    \vspace{-3mm}
\end{figure}

Embodied AI requires an agent to have an understanding of the environment and to update its knowledge based on new observations. 
In this paper, our goal is to model scenes with our Continuous Scene Representations (\csr), which can then be used in downstream visual and embodied tasks (Fig.~\ref{fig:overview}).
We start by giving an overview of \csr (Sec. \ref{sec:overview}). Then we address learning and updating the scene representation (Sec. \ref{sec:representation}). 
Finally, we describe strategies for linking \csr to agent embodiment (Sec. \ref{sec:grounding}) and detecting changes in the scene (Sec.~\ref{sec:changes}). 

\subsection{Overview}
\label{sec:overview}
Given a navigating agent in a scene (e.g., a robot in a kitchen), our goal is to create a \csr of the room. 
A \csr is a graph where nodes represent objects and edges represent the relationships between them.
Both nodes and edges are represented by continuous valued features.
As the agent moves, the \csr updates to accommodate observed objects and relationships.
The agent is situated within the scene representation to allow for planning.

\noindent\textbf{Features for nodes and edges.}
To motivate continuous valued nodes and edges, consider a scene graph that models the \textsc{[support]} relation.
In a kitchen scene, the graph might include a directed edge from \textsc{[stove]} to \textsc{[kettle]}, representing the ``kettle is on the stove". This representation throws away useful information like the appearance of the relationship or the relative positions of the objects. If instead, the representation was a continuous vector, it could preserve more information. Moreover, a continuous representation can implicitly encode many other relations (such as \textsc{[next-to]} and \textsc{[inside]}), which could be recovered when necessary using light-weight, trainable linear projection heads. Hence, learning continuous features for relations could provide a flexible representation that could conceivably adapt based on an agent's needs. 

\noindent\textbf{Scene representation consistency.}
To motivate consistency of the representation, consider again the example of the kettle.
As an agent navigates in the scene, it should determine if a kettle is the same object from previous timesteps to accurately model the underlying relationships.
In the terminology of \csr, as the agent moves, it must decide if observed objects match with existing nodes or should spawn new nodes.
Edges should be propagated accordingly.
This is challenging, especially in our setting that uses only RGB images as input.

\noindent\textbf{Adding the embodiment.}
\csr does not directly model the agent or its actions.
Such information must be captured if an agent is to navigate to a particular \csr node.
For example, given a node corresponding to \textsc{[kettle]}, it would be useful to retrieve the sequence of actions to get to a position where the kettle is observable.

\noindent\textbf{Scene representation notation.}
A \csr is a graph $\mathcal{G_{\text{\csr}}(\mathcal{N_{\text{\csr}}}, \mathcal{E_{\text{\csr}}})}$, where $\mathcal{N_{\text{\csr}}}, \mathcal{E_{\text{\csr}}}$ are the sets of nodes and edges, respectively. Each node and edge is a vector in $\mathbb{R}^L$, $L$ being the feature dimension. A local scene representation $\mathcal{G}^t_{\text{\csr}}$ is the continuous scene representation constructed based on the observation of the agent at time $t$.
There is a detection function that takes $I_t \in \mathbb{R}^{H \times W \times 3}$ and extracts bounding box masks $\{r_1, r_2, ..., r_n \}$, $r_i \in \{0, 1\}^{H \times W}$, one for each object.
Each mask $r$ has value one within the box, else zero.
A \csr encoder $f_\text{\csr} (I_t, r_i, r_j) \mapsto y_{i, j} \in \mathbb{R}^L$ is introduced to map directed pairs of objects to their feature representations, where $ y_{i, j}$ is the continuous valued vector representing the feature between node $i$ and $j$.
If $i=j$, $y_{i, j} \in \mathcal{N_{\text{\csr}}}$, otherwise $y_{i, j} \in \mathcal{E_{\text{\csr}}}$.
A matching function $M_{\text{\csr}}$, aggregates the local observation into $\mathcal{G_{\text{\csr}}}$ by taking an existing \csr and a local scene representation $\mathcal{G}^t_{\text{\csr}}$ and updating $\mathcal{G_{\text{\csr}}}$.

\noindent\textbf{Embodied state graph notation.}
An embodied state graph $\mathcal{G_{\text{state}}(\mathcal{S}, \mathcal{T})}$,
contains states $\mathcal{S}$ pointing to nodes in the \csr, with one state for every egocentric observation, and action transitions $\mathcal{T}$ between states.
The nodes $\mathcal{S}$ keep track of references to the objects and relationships in $\mathcal{G_{\text{\csr}}}$ that are seen by the agent at that state.
A transition takes place via agent actions (e.g., \textsc{MoveForward}).
Hence, $\mathcal{G_{\text{state}}}$ stores information about how the agent should reach particular visual observations.

\subsection{Creating the Scene Representation}
\label{sec:representation}
In this section, we explain how to create a \csr or $\mathcal{G_{\text{\csr}}}$, which are used interchangeably.
We first discuss building a local representation from the current egocentric view and learning node and edge representations.
Since the agent explores over time, we then describe how the local information is aggregated.

\begin{figure}[tp]
    \centering
    \includegraphics[width=\linewidth]{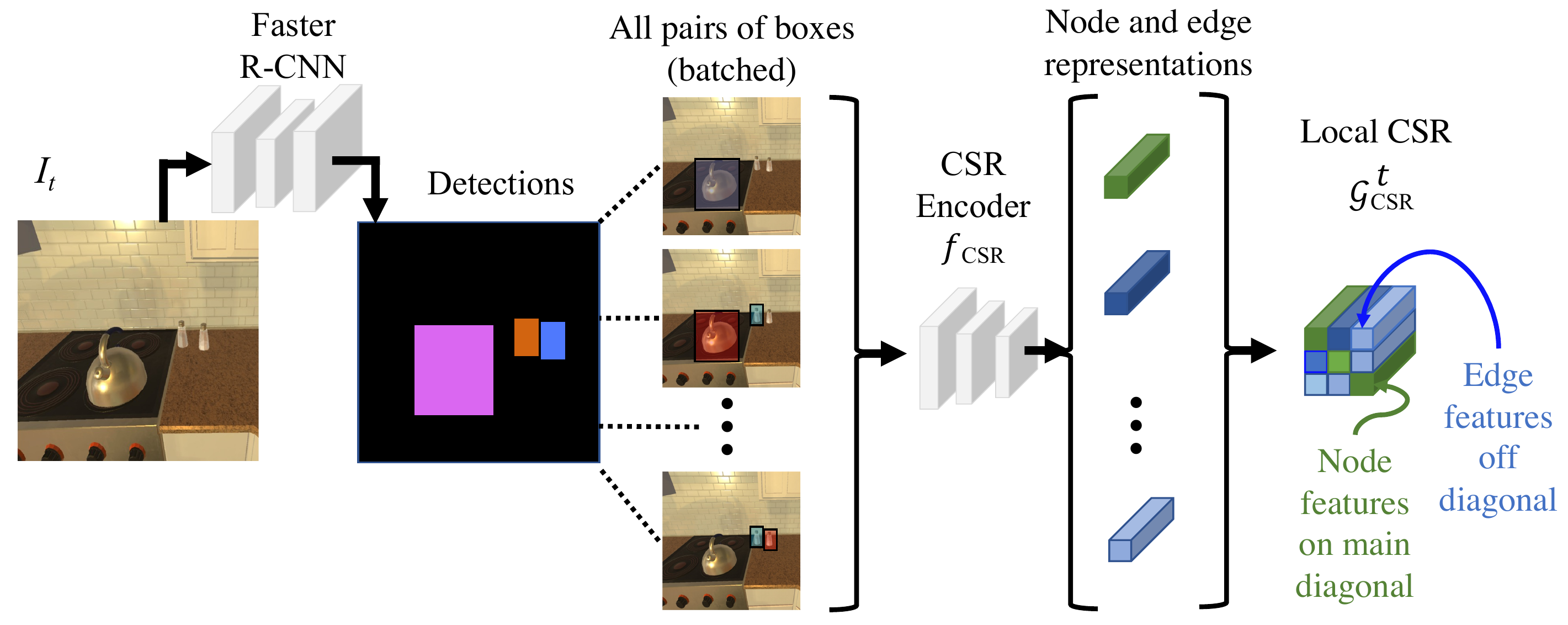}
    \caption{
        \textbf{Building the Local Graph.}
        A detector extracts $n$ boxes and all $n^2$ pairs of of boxes are batched and fed to the \csr encoder. This yields the \csr representation for this image. Node features are in green and edge features are in blue.
    }
    \label{fig:local_graph}
    \vspace{-3mm}
\end{figure}

\noindent\textbf{Building the local scene representation.}
In conventional scene representations, nodes are represented by class labels (e.g., \textsc{[mug]} and \textsc{[sink]}), and edges by attributes relating classes (e.g., \textsc{[under]} and \textsc{[inside]}).
Instead \csr uses continuous features for nodes and edges.
To get box region proposals corresponding to nodes, a detector is employed on the current image $I_t$, which yields $n$ detections.
In practice we use a Faster-RCNN model; however, class labels are discarded.
All $n^2$ pairs of regions are mapped into a latent space, using the \csr encoder $f_{\text{\csr}}$ as depicted in Fig. \ref{fig:local_graph}.
A batch of $n^2$ items is fed through $f_{\text{\csr}}$, with each item in the batch containing (1) a unique ordered pair of regions and (2) $I_t$.
The result is $n^2$ feature vectors, which form the local scene representation $\mathcal{G}^t_{\text{\csr}}$ with $n$ node features and $n^2-n$ edge features.
While \csr feature extraction is quadratic in the number of detections, all computations are independent and hence parallelizable.

What is a good objective to learn the \csr encoder $f_{\text{\csr}}$?
Our high-level idea is to embed the same relationship from different view points to the same location in the embedding space, assuming objects do not move within a trajectory.
Consider again the kettle on the stove.
The relationship of the kettle to the stove can be observed from different angles, with a variety of objects in the background.
However, the global characteristics of the relationship do not change across views.
The same argument holds for nodes.

\begin{figure}[tp]
    \centering
    \includegraphics[width=19pc]{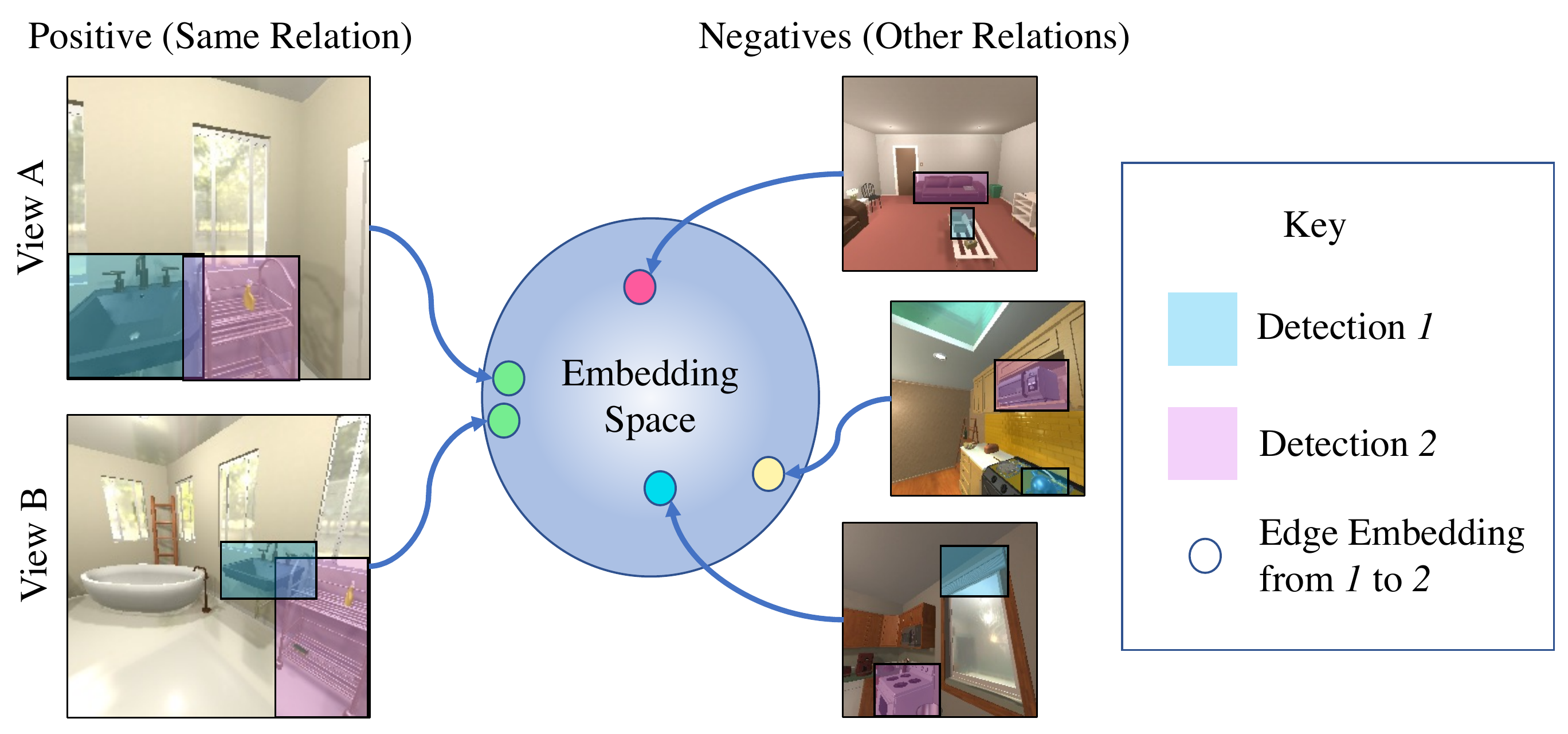}
    \caption{
        \textbf{Embedding Space.}
        The loss encourages the same relationship observed from different views to be close in feature space and far from other relationships.
    }
    \label{fig:loss}
    \vspace{-3mm}
\end{figure}

To learn features capable of registering relations from different viewpoints, we take inspiration from the momentum contrast approaches \cite{he2020momentum,chen2020improved}. 
Given two views of the same pair of objects, a representation is extracted from the first view via $f_{\text{\csr}}$ and from the second view via a momentum encoder, which has the same architecture and initialization as $f_{\text{\csr}}$, but is updated via a slow moving average of $f_{\text{\csr}}$ weights instead of by gradient updates.
InfoNCE loss \cite{oord2018representation} is used to enforce similarity of relations in feature space,
$
\label{infonce}
    \mathcal{L}_{\text{\csr}} = -\log\frac{\exp({\mathsf{CosSim}(q, k^+)/\tau})}{\sum_{i=0}^K\exp({\mathsf{CosSim}(q, k^i)/\tau})}.
$
Here $q$ is the feature representation of a node or edge outputted by $f_{\text{\csr}}$, $k^+$ is the momentum encoder representation for the second view of the same relationship, each $k^i$ is a negative drawn from a bank of $K$ other momentum encoded relationships, $\tau$ is a softmax temperature scaling parameter, and $\mathsf{CosSim}$ is the dot product between L2 normalized features.
Notably this objective encourages learning features that (1) have high $\mathsf{CosSim}$ when they represent that same relationship, (2) are multi-view consistent in the feature space, and (3) are different from those of other relationships. See Fig.~\ref{fig:loss} for visualization of these attributes.

\noindent\textbf{From local to global scene representation.}
Considering the agent is navigating over time, it is not sufficient to capture information from a single view.
As the agent traverses, it is necessary to determine if node representations are new objects or correspond to previously seen objects.
To handle such complexity, we design a way to update nodes and edges based on the current local observation.

\begin{figure}[tp]
    \centering
    \includegraphics[width=\linewidth]{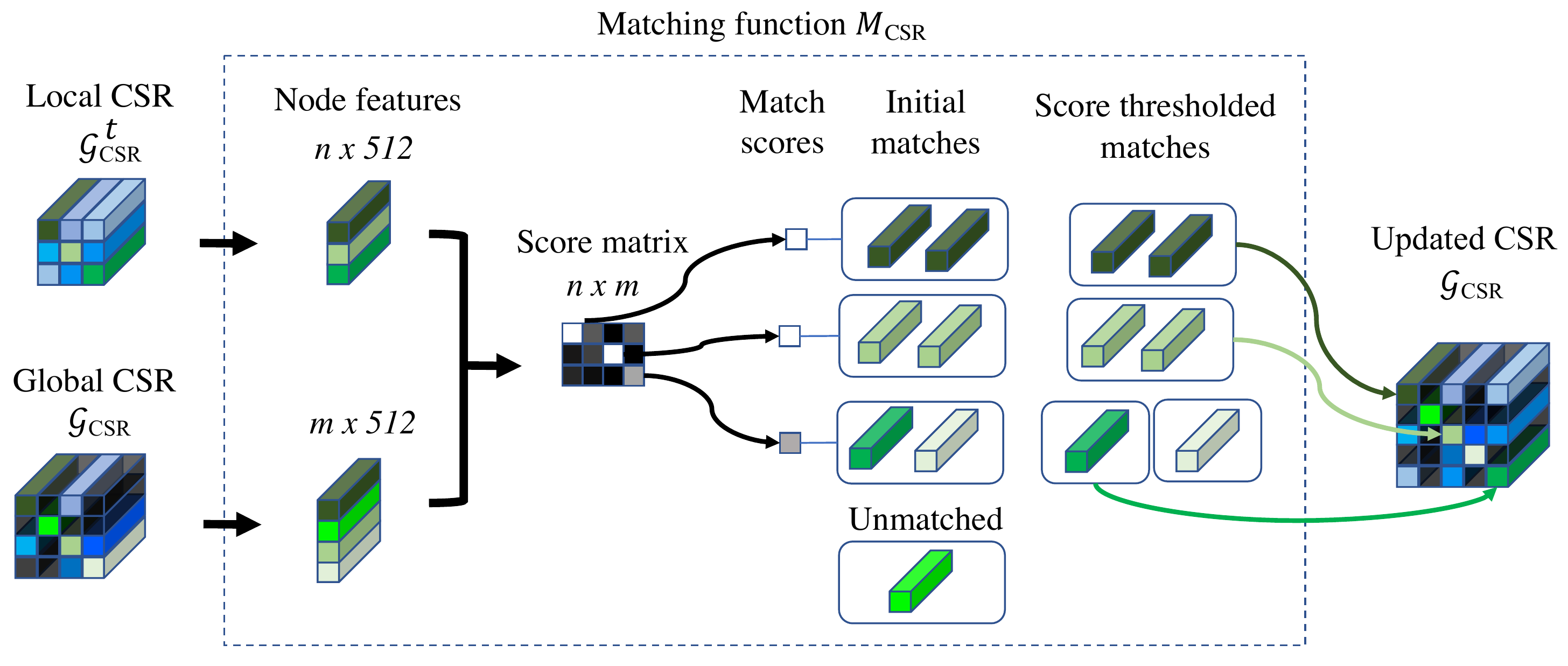}
    \vspace{-0.4cm}
    \caption{
        \textbf{Matching.}
        A function $M_{\text{\csr}}$ takes the local and the existing global \csr to update the global \csr.
    }
    \label{fig:matching}
    \vspace{-4mm}
\end{figure}

Our insight is to use a matching function $M_{\text{\csr}}$ between the node embeddings from the current local scene representation $\mathcal{G}^t_{\text{\csr}}$ and $\mathcal{G_{\text{\csr}}}$ to update $\mathcal{G_{\text{\csr}}}$.
At the first timestep $\mathcal{G}_{\text{\csr}} \gets \mathcal{G}^t_{\text{\csr}}$.
For subsequent timesteps, node features corresponding to the same object should have high $\mathsf{CosSim}$. As shown in Fig. \ref{fig:matching}, $M_{\text{\csr}}$ first computes $\mathsf{CosSim}$ between pairs of features in $\mathcal{N}_{\text{\csr}}$ and $\mathcal{N}^t_{\text{\csr}}$, which yields a score matrix.
Next a maximal linear score assignment (Hungarian match) creates an initial matching of nodes.
If the score of a match exceeds a threshold, the nodes are considered a true match (more details in Sec.~\ref{sec:rearrangement}).
The matched local node features are then averaged with their matches to update the \csr nodes.
For edges, if a new relationship is observed it is added otherwise it is averaged into the representation in an analogous fashion to the nodes.
Unmatched nodes and corresponding edges from $\mathcal{G}^t_{\text{\csr}}$ are then directly added as new nodes and relations.

\subsection{Linking \csr to Agent States for Planning}
\label{sec:grounding}
We have discussed how to create our representation, which captures object and their relationships in the scene.
How should this representation be used to solve embodied tasks?
While using the representation as additional input for downstream training is an option, we present a simple alternative.
Namely, we discuss the creation of an embodied state graph $\mathcal{G_{\text{state}}}$ to enable planning within the \csr.
As the agent traverses, we keep track of (1) scene nodes that are observed in each egocentric view, which initialize states in $\mathcal{G_{\text{state}}}$ and (2) the actions that lead to changes in views, which initialize the transitions in $\mathcal{G_{\text{state}}}$.
To navigate to a specific scene feature in $\mathcal{G_{\text{\csr}}}$, the agent can lookup a corresponding state where this feature was observed, and navigate to it by planning actions from its current state to the target state in $\mathcal{G_{\text{state}}}$.
Since the states in $\mathcal{G_{\text{state}}}$ are linked via actions, the path and actions to the target object can be recovered using a standard breath first search planner over the state graph. Note, this is a simple planning strategy, is independent of \csr creation.
In the presence of noise, it is also possible to employ more sophisticated closed-loop planning.

\subsection{Detecting Changes}
\vspace{-0.2cm}
\label{sec:changes}

In practice, scenes typically change over time. 
For example, kitchen equipment frequently moves.
Given two trajectories, each observing a different static configuration, it is important to determine the objects that changed.

\noindent\textbf{Motivating between-trajectory object correspondences.}
Our insight is to detect changes in the scene by leveraging object correspondences between trajectories.
If a kettle is observed on a stove in one trajectory and on a countertop in another, the agent should know that plausibly this kettle has moved.
While our scene representation captures nodes and edges in the context of the scene, we would also like to detect objects that shuffle over time.
This motivates finding object correspondences between trajectories.

\noindent\textbf{Between-trajectory object features.}
Our idea is to augment the scene nodes of $\mathcal{G_{\text{\csr}}}$ with a object feature vector that is responsible for matching the object instance between trajectories.
This feature is learned similarly to the scene node features.
However, instead of positives for the contrastive loss coming from different views in the same static scene, they come from different views across varied initializations of the scene.
The representation learner cannot rely on consistent scene information to minimize the low, but should instead focus only on embedding characteristics of the object instances.
Using a similar Hungarian matching scheme, as in Sec.~\ref{sec:representation}, object features are matched between the two trajectories and thresholded based on $\mathsf{CosSim}$.

\noindent\textbf{Using low feature similarity to detect change.}
Recall, when learning node and edge features, we leveraged that the scene was static.
Based on the object matches, detecting violations to the static assumption provides signal for what has moved.
Concretely, using the object feature assignments, the $\mathsf{CosSim}$ between matched scene nodes and edges (i.e., scene features before object features were introduced) is computed.
If scores are lower than a threshold there is signal the visual object has moved (i.e., the underlying spatial relationship has changed).
In summary, the object features are used to match object instances between trajectories and then low scene node/edge feature similarity is used to determine what moved.

\section{Experiments}
\label{sec:experiments}
In this section, we first present our datasets (Sec.~\ref{sec:datasets}).
We show competitive performance of Continuous Scene Representations (\csr) on 2-phase visual room rearrangement~\cite{weihs2021visual,batra2020rearrangement}, \emph{without training for the task} (Sec.~\ref{sec:rearrangement}), \csr encodes spatial information that can be recovered via linear probes (Sec.~\ref{sec:probe}), \csr can retrieve layout-consistent views of the scene against hard negatives (Sec.~\ref{sec:change}), and \csr is effective for tracking objects in real-world data (Sec.~\ref{sec:ycb}).

\subsection{Datasets}
\vspace{-0.2cm}
\label{sec:datasets}

For rearrangement experiments (Sec.~\ref{sec:rearrangement}), we use the RoomR dataset introduced by Weihs \textit{et al.} \cite{weihs2021visual}.
Its test and val sets each contain 1000 room rearrangements.

To train the \csr encoder $f_{\text{\csr}}$, which embeds nodes and edges, we generate a dataset using AI2-THOR~\cite{ai2thor}.
For each of the 80 train rooms, we render 20 random agent poses in 5 different room initialization with random lighting, texture, and object locations.
RGB images for this and all following datasets are $224\times224$.
We mine pairs of objects (i.e., edges) that are viewed from different view points in each room initialization to get a dataset of 900K samples.

We additionally create smaller datasets for our various experiments.
We create two image datasets of $\sim$5k images in 80/20 train/test splits with spatial relationship labels for pairs of ground truth boxes (e.g., ``$x$ supports $y$").
These are used in linear probes for our edge representations (Sec. \ref{sec:probe}). 
A test dataset of 4.4K triplets of images is generated, where a query and a positive image have the same static scene layout and a negative image presents the scene with one object moved.
We use this set to evaluate the model's matching ability to retrieve the positive against a hard negative (Sec. \ref{sec:change}).
Finally we use the YCB-Video test set \cite{xiang2018posecnn} of 12 videos to evaluate the tracking and updating components of our method on real world data (Sec. \ref{sec:ycb}).

Across all datasets, train, validation, and test rooms are constant.
More details on the Faster-RCNN detector and the exploration policy network can be found in Appx.~\ref{appx:rcnn}, \ref{appx:policy}.

\subsection{Downstream Rearrangement}
\vspace{-0.2cm}
\label{sec:rearrangement}
The goal is to evaluate \csr for a downstream embodied task, 2-phase visual room rearrangement~\cite{weihs2021visual}.
Rearrangement serves as a difficult, but practical, end-task with applications to indoor service robots.
Furthermore, performing visual room rearrangement requires capable perception, memory, navigation, and planning, thereby testing how the various components of our approach work together.

\begin{table*}
  \small
  \centering
  \tabcolsep=0.25cm
  \begin{tabular}{l?cccc?cc?cc?cc}         
  \toprule
                          &\multicolumn{4}{c?}{\textsc{Inferred}} & \multicolumn{2}{c?}{100 $\cdot$ \textsc{Success} $\uparrow$} & \multicolumn{2}{c?}{100 $\cdot$ \textsc{\% FixedStrict} $\uparrow$} & \multicolumn{2}{c}{\textsc{\%E} $\downarrow$}\\
Experiment                &plan & match & box & traj.& Val.  & Test & Val. & Test & Val.  & Test\\\midrule

\csr (GT-MBT)      &\cmark&\xmark&\xmark&\xmark& 8.8   & 10.0  & 26.0  & 27.0   & 0.74  & 0.73\\
\csr (GT-BT)                &\cmark&\cmark&\xmark&\xmark& 3.0   & 2.2  & 7.9  & 5.9   & 0.97  & 0.98\\
\csr (GT-T)           &\cmark&\cmark&\cmark&\xmark& 1.3   & 0.7  & 3.8  & 2.1   & 1.07  & 1.11\\\midrule
VRR \cite{weihs2021visual} &-&-&-&-& 0.5  & 0.2  & 1.2  & 0.7   & 1.15 & 1.12\\
VRR+Map \cite{weihs2021visual} &-&-&-&-& 0.6  & 0.3 & 1.6  & 1.4 & 1.15 & \textbf{1.10}\\
\csr (Ours)              &\cmark&\cmark&\cmark&\cmark& \textbf{1.2}   & \textbf{0.4}  & \textbf{3.3}  & \textbf{1.9}   & \textbf{1.13}  & 1.17\\
\bottomrule
  \end{tabular}
  \vspace{-0.3cm}
  \caption{ \textbf{Downstream Rearrangement.} Using \csr, we are able to outperform the RL based visual room rearrangement baselines in the hard 2-phase setting (i.e., walkthrough and unshuffle phases happen sequentially).
  Additionally, the improved performance when including ground truth components, hints that as the various components of our approach improve, our end-to-end solution will also improve.
  }\label{tab:exp_rearrangement}
  \vspace{-0.2cm}
\end{table*}

\noindent\textbf{Task.}
In a walkthrough phase, an agent observes objects in their target configurations, building a \csr and embodied state graph: $(^{\text{walk}} \mathcal{G_{\text{\csr}}}, ^{\text{walk}} \mathcal{G_{\text{state}}})$.
The agent leaves the room and 1-5 objects change (e.g., a mug moves from a countertop to a table).
In the unshuffle phase, the agent returns and must restore the room to its original layout.
In our case, a second unshuffle \csr and embodied state graph: $(^{\text{un}} \mathcal{G_{\text{\csr}}}, ^{\text{un}} \mathcal{G_{\text{state}}})$ is built.
Leveraging the fact the agent enters the room from the same starting location, we fuse the $^{\text{walk}} \mathcal{G_{\text{state}}}$ and $^{\text{un}} \mathcal{G_{\text{state}}}$ into a unified embodied state graph $^{(\text{walk},\text{un})} \mathcal{G_{\text{state}}}$, using the initial state as a point of correspondence between two trajectories.
While this simplifying assumption is allowable due to the task construction of \cite{weihs2021visual}, it is plausible that state correspondence can be established using features, which is left to future work.

By identifying target objects that moved between creation of $^{\text{walk}} \mathcal{G_{\text{\csr}}}$ and $^{\text{un}} \mathcal{G_{\text{\csr}}}$ (i.e., high object feature similarity but low node feature similarity as discussed in Sec.~\ref{sec:changes}), it is possible to query current and target locations for the moved objects in $^{(\text{walk},\text{un})} \mathcal{G_{\text{state}}}$.
Using a simple breath first search planner over $^{(\text{walk},\text{un})} \mathcal{G_{\text{state}}}$, the agent plans to the location of the target object, executes a pick-up action, and navigates to the goal location, where the object is placed.

\noindent\textbf{Model.}
We use an exploration agent to explore the room during the two phases of the task.
We use a simple exploration policy network with three convolutional layers and a GRU, which takes RGB images and outputs a distribution over navigation actions $\mathcal{A}_\text{nav} =$ \{\textsc{MoveForward, MoveLeft, MoveRight, MoveBack, RotateRight, RotateLeft, LookUp, LookDown}\}.
The network is trained using DD-PPO \cite{Schulman2017ProximalPO,Wijmans2020DDPPOLN} in the AllenAct framework \cite{Weihs2020AllenActAF}, with a reward that encourages visiting more states and objects.
Note: this objective has no notion of rearrangement.
To generate the object region proposals used to generate box mask inputs for the \csr encoder, we use a Faster-RCNN~\cite{ren2015faster} detector.
This detector is trained on LVIS \cite{Gupta2019LVISAD} and AI2-THOR data.
Leveraging the policy network and detector during walkthrough and unshuffle, we get \csr representations for each trajectory, with corresponding embodied state graphs.
We then detect changes in the scene and execute a plan to restore objects.
For more details on the exploration training, reward, detector training, loss, and hyperparameters, see Appx.~\ref{appx:rcnn}, \ref{appx:policy}, and \ref{appx:csr}.

\noindent\textbf{Metrics.}
We adopt rearrangement metrics \cite{weihs2021visual}:
\begin{itemize}[leftmargin=*]
\itemsep-0.1em
    \item \textsc{Success}: The percentage of episodes in which \emph{all} the objects are correctly returned to their original states.
    \item \textsc{\%FixedStrict}: The percentage of objects returned to their proper place over all rooms. However, if a mistake is made in a room (e.g., an object is moved that should not be), the value is 0 for that rearrangement even if other objects are returned correctly.
    \item \textsc{\% E}: A ratio between the energy of a room before and after rearrangement. The energy function decreases monotonically as the agent brings objects closer to their target destinations. A value of 0 corresponds to a perfect rearrangement, 1 to no change, and greater than 1 to a final configuration further from the goal than the initialization.
\end{itemize}

\noindent\textbf{Ablation.}
We ablate \csr in the context of the rearrangement task.
We substitute ground truth (GT) for key components of our model to ablate the impact of their performances on the rearrangement task.
We study the effects of using GT \textbf{M}atching, \textbf{B}oxes, and \textbf{T}rajectories.
For GT matching, we substitute the matching function $M_{\text{\csr}}$ for GT object instance labels, which allow for perfect matching, both within a trajectory and between walkthrough and unshuffle trajectories.
For GT boxes, we substitute the Faster-RCNN detector for GT boxes.
For GT trajectories, we substitute the trained exploration policy for a heuristic policy that visits all objects that get shuffled in both walkthrough and unshuffle phases.
Concretely, we compare against the following ablated models:
\begin{itemize}[leftmargin=*]
\itemsep-0.2em
\item \textit{\csr (GT-MBT)}: GT object \textbf{M}atches, \textbf{B}oxes, and heuristic \textbf{T}rajectories.
\item \textit{\csr (GT-BT)}: Matching is estimated within a trajectory and between the two trajectories, keeping box and trajectory GT constant from \csr (GT-MBT).
\item \textit{\csr (GT-T)}: Boxes are estimated using the Faster-RCNN detector, keeping the heuristic exploration trajectories constant from \csr (GT-BT).
\end{itemize}

\noindent\textbf{Baselines.}
We compare against the following prior work for the challenging 2-phase rearrangement setting:
\begin{itemize}[leftmargin=*]
\itemsep-0.2em
\item \textit{VRR}: Corresponds to the (RN18, PPO+IL) from \cite{weihs2021visual}.
The model uses a frozen ImageNet pretrained ResNet-18 and a mixture of imitation learning and reinforcement learning to train \emph{directly} on the rearrangement task.

\item \textit{VRR+Map}: Simmilar to VRR, but with the addition of Active Neural SLAM \cite{chaplot2020neural} for mapping. Corresponds to (RN18+ANM,PPO+IL) from~\cite{weihs2021visual}.

\end{itemize}

\noindent\textbf{Is \csr competitive v.s. sophisticated task-specific baselines?}
The results in Tab.~\ref{tab:exp_rearrangement} support that using a simple exploration policy and detector in concert with \csr performs favorably on the \textsc{Success} and \textsc{\%FixedStrict} metrics compared to the sophisticated baselines trained directly on the task.
This shows the effectiveness of \csr since  \textit{no part of our approach was trained to rearrange rooms}.

\noindent\textbf{How do improved submodules improve the solution?}
We replace each \csr submodule with GT as a proxy for an improved submodule.
See Tab.~\ref{tab:exp_rearrangement}.
Looking at \csr (GT-T), we see that performance is comparable to \csr (Ours), suggesting that the trained exploration agent generalizes sufficiently.
Comparing \csr (GT-BT), we see larger gains suggesting that with better detection, performance would improve by $\sim 3$ times.
Finally when comparing to \csr (GT-MBT), we find that there is room for improvement in matching.
These results suggest a high ceiling to our pipeline.

\subsection{Representation Linear Probes}
\label{sec:probe}

We probe if \csr encodes interpretable relationships.

\noindent\textbf{Tasks.}
We linearly probe \csr features for two relationship classification tasks. 
First, given a pair of box regions of interest, predict if one object supports the other. We call this the \textsc{[support]} task. This is a three-way classification problem, where object 1 either supports object 2, vice versa, or there is no support relationship.
The second task is predicting, given a pair of boxes, if they are supported by the same surface.
We call this the \textsc{[sibling]} task.
In this binary task, the direction of the edge does not matter.

\noindent\textbf{Model.}
We fit a linear layer on a frozen \csr feature extractor $f_{\text{\csr}}$.
We train two linear heads for \textsc{[support]} and \textsc{[sibling]} respectively.

\noindent\textbf{Metrics.}
We consider the mean accuracy across both tasks.

\noindent\textbf{Baselines.}
We compare to the following baselines:
\vspace{-1mm}
\begin{itemize}[leftmargin=*]
\itemsep-0.3em
\item \textit{Random Features}: Randomly initialized and frozen ResNet-18 receiving 5 channels as input (3 for RGB and 2 for object boxes).
\item \textit{Supervised \textsc{[support]}}: a ResNet-18 backbone trained end-to-end on the \textsc{[support]} task. The representation is frozen and a separate linear head is trained for the \textsc{[sibling]} task.
\item \textit{Supervised \textsc{[sibling]}}: Similar to the \textit{Supervised \textsc{[support]}} model but trained end-to-end on the \textsc{sibling} task.
\item \textit{Fine-tuned \textsc{[support]}, Fine-tuned \textsc{[sibling]}}: Same as \textit{Supervised \textsc{support}, \textsc{sibling}}; however, encoders are initialized with \csr contrastively pre-trained weights instead of random initialization.
% \item \textit{Fine-tuned \textsc{sibling}}: Same as \textit{Supervised \textsc{sibling}}, again with \csr init.
% \vspace{-5mm}
\end{itemize}

\begin{table}
 \scriptsize 
  \centering
  \tabcolsep=0.3cm
  \begin{tabular}{l?c|c}
  \toprule
Experiment              & Linear Probe mAcc & Retrieval Acc.\\\midrule
Random Chance           & 41.7 & 50.0\\
Random Features         & 51.8 & 52.4\\
Supervised \textsc{[support]}   & 84.3 & 69.3\\
Supervised \textsc{[sibling]}      & 70.2 & 54.9\\
Fine-tuned \textsc{[support]}   & 77.7 & 64.6\\
Fine-tuned \textsc{[sibling]}      & 78.3 & 57.6\\
\midrule
\csr features           & \textbf{85.3} & \textbf{84.3}\\
\bottomrule
  \end{tabular}
  \vspace{-0.3cm}
  \caption{ \textbf{Linear Probes and Retrieval.}
  \csr features outperform supervised and fine-tuned baseline features.
  }
  \label{tab:exp_rel_probe}
  \vspace{-3mm}
\end{table}

\noindent\textbf{Benefits to continuous instead of discrete relations.}
While the baselines outperforms \csr for the specific tasks they are trained for, when averaging performance over both tasks, \csr outperforms the baselines, as seen in Tab.~\ref{tab:exp_rel_probe} (left).
These results support that training on specific relations can be detrimental in transfer setting and that our features can be used to recover specific spatial relations.

\subsection{Retrieving Consistent Relationships}
\label{sec:change}
Our \csr learning objective assumes scenes are static, enforcing the similarity of relations between objects seen from different viewpoints.
We hypothesize that our features should allow for detecting violations to this assumption.

\begin{figure}[tp]
    \centering
    \includegraphics[width=\linewidth]{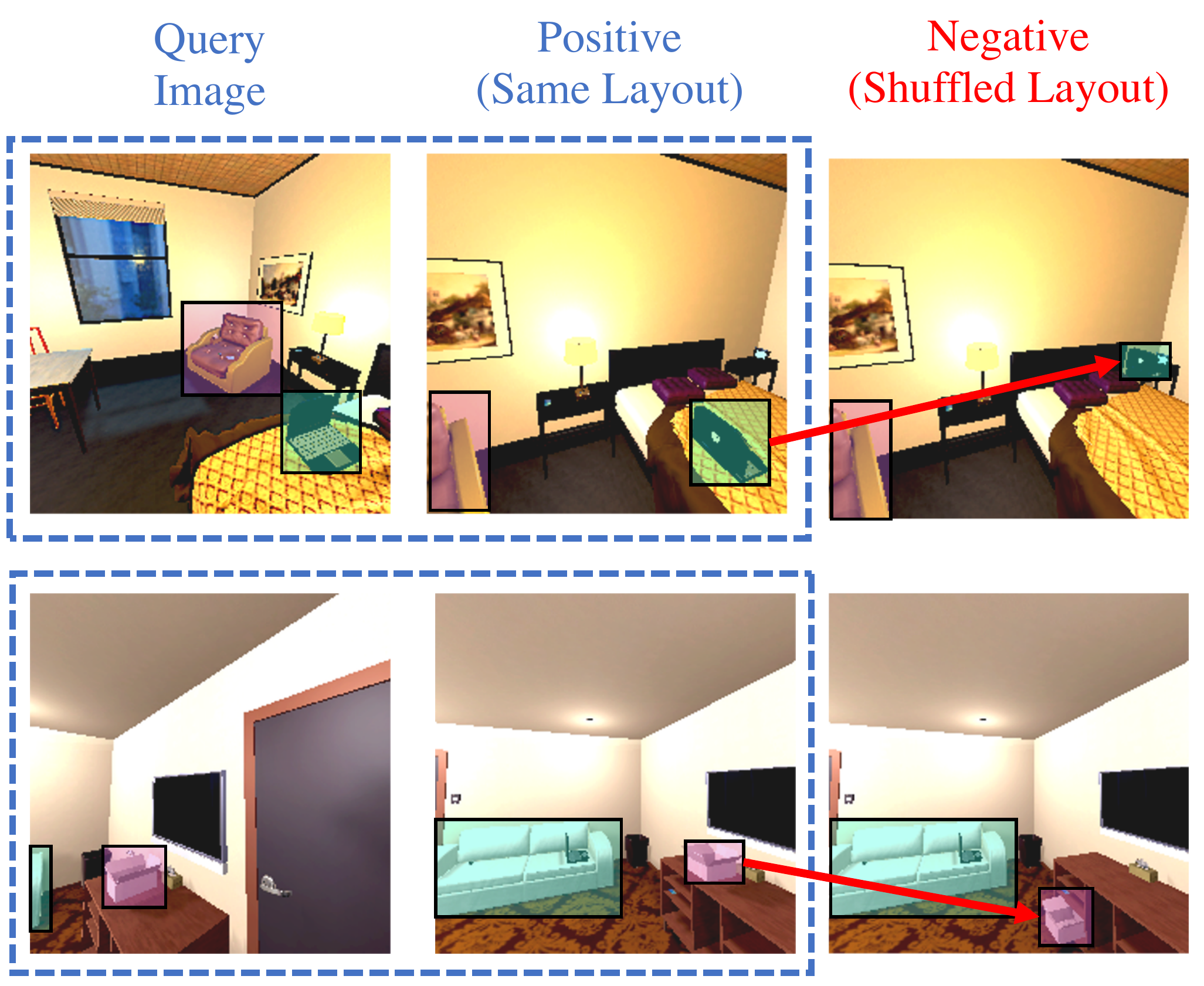}
    \vspace{-0.5cm}
    \caption{
        \textbf{Consistent relationship sample data.}
        Features for edges in the same room layout should be more similar than features in different layouts even with relatively small pixel differences in the positive and negative images.
    }
    \label{fig:retrival}
    % \vspace{-5mm}
    \vspace{-0.3cm}
\end{figure}

\noindent\textbf{Task.}
Consider triplets of images.
The first is a query image containing a pair of objects.
The second is the \emph{positive} image taken from a different vantage point that also captures the same pair.
The third  is the \emph{negative} image taken from the same pose as the second image; however, one of the objects in the pair is moved.
The task then is to match the query and the positive image with $\mathsf{CosSim}$ of features as the matching criteria.
Fig.~\ref{fig:retrival} shows sample data.

\noindent\textbf{Model.}
We use our $f_{\text{\csr}}$ directly to embed three relationships.
We compute the $\mathsf{CosSim}$ between the query features against the features from the positive and the negative images.
To solve the classification task, we choose the image that has the higher $\mathsf{CosSim}$ score.

\noindent\textbf{Metrics.}
We measure accuracy of the predictions.

\noindent\textbf{Baselines.}
We adopt the same baselines as in Sec.~\ref{sec:probe}. The Supervised \textsc{[support]} baseline is the main point of comparison. This model determines if two objects participate in a support relationship (i.e., one on top of the other). Because changes in the dataset involve objects shifting to different supporting surfaces, these events should correspond to drastic changes in feature space.

\noindent\textbf{Results.}
Based on the results in Tab.~\ref{tab:exp_rel_probe} (right), we find that \csr outperforms the baselines in terms of matching the query and positive images.
See Fig.~\ref{fig:retrival}, which qualitatively illustrates the difficulty of the task.
The results suggest that \csr learns some underlying spatial relationships.

\subsection{Real World Tracking}
\vspace{-0.2cm}
\label{sec:ycb}

To demonstrate the potential of the representation on real-world data, we additionally evaluate \csr features, trained in simulation
on object tracking in YCB-Video~\cite{xiang2018posecnn}.

\noindent\textbf{Task.}
The goal is to track objects over a trajectory. We formulate the problem as clustering, where, with each new observation, a detected object can either be assigned to an existing cluster (representing a previously observed object) or to a new cluster.
The target GT clustering is one cluster for each object instance, with a count per cluster corresponding to the number of times the object was detected.

\noindent\textbf{Model.} \csr feature extraction and matching within a trajectory. Note, we use GT boxes for detections in lieu of a detector. Each node represents a cluster with number of elements equal to the number of assignments to that cluster.

\noindent\textbf{Metrics.}
For each video, we measure the Adjusted Rand Index (ARI)~\cite{hubert1985comparing}, a common clustering accuracy metric, between the \csr and GT assignments.
A value of 1 corresponds to perfect clustering and 0 to random clustering.

\noindent\textbf{Ablations/Baselines:}
\vspace{-0.3cm}
\begin{itemize}[leftmargin=*]
\itemsep-0.2em
    \item \textit{No-Update}: Node features are fixed at initialization without the feature updating discussed in Sec.~\ref{sec:representation}.
    \item \textit{Opt.}: Oracle with feature updating and matching threshold set optimally.
\end{itemize}

\begin{figure}[tp]
    \centering
    \includegraphics[width=\linewidth]{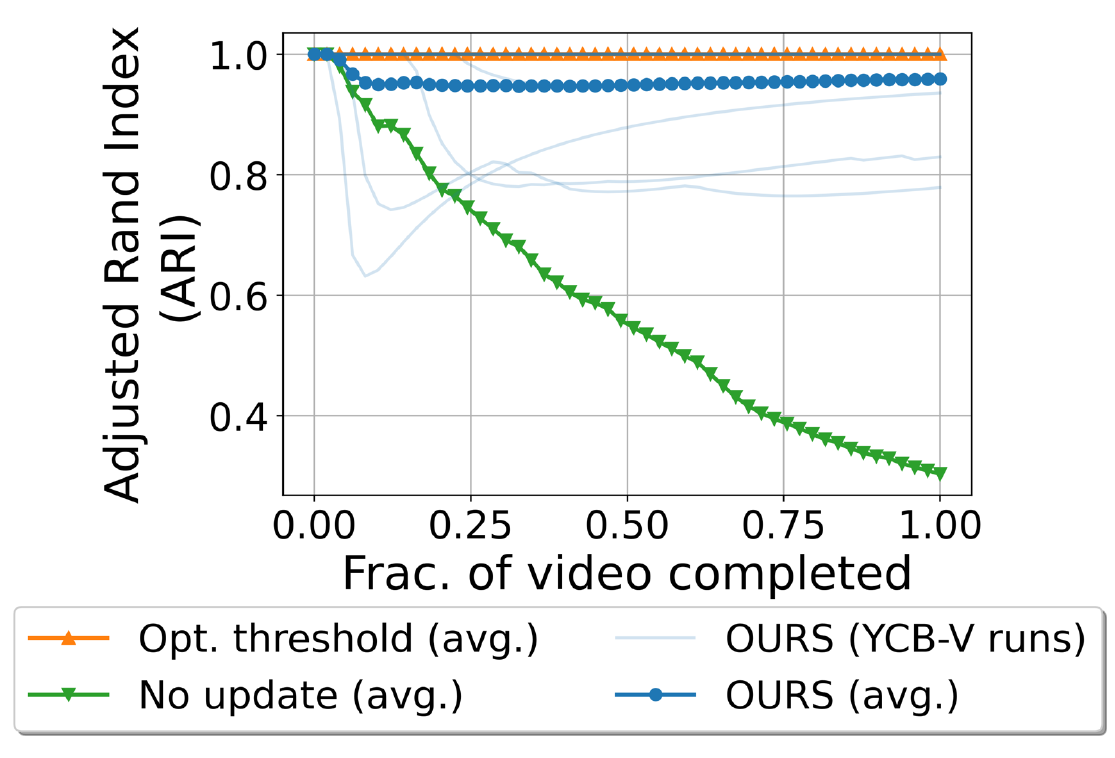}
    \vspace{-2em}
    \caption{
        \textbf{Matching on YCB-Video.}. Tracking performance is close to optimal and outperforms an ablated version without feature updating.
    }
    \label{fig:real_world}
    \vspace{-0.3cm}
\end{figure}

\noindent\textbf{Results.}
In Fig.~\ref{fig:real_world} we see near perfect matching (compared to the baseline \textit{Opt.}) and the usefulness of updating features (compared to \textit{No-Update} baseline).
These preliminary results suggest that \csr trained in simulation can be applied to real-world data distributions and that updating the representation on-the-fly is critical for tracking.

\section{Limitations}
\vspace{-0.2cm}
In this work, we mainly focus on capturing the pose of objects and ignore the state of the objects (e.g., open v.s. closed microwave). The state of objects plays an important role in task planning. Future work will consider incorporating the state of the object. Also, the interaction of the agents with the objects is not used to update the scene representations. Hence, future work will address agent-object interactions. 
Additionally developing learned matching functions, which go beyond cosine similarity, is a promising direction to develop better matching performance.

\vspace{-0.2cm}
\section{Conclusion}
\vspace{-0.2cm}
We present Continuous Scene Representations (\csr) to model objects and their relations as continuous valued feature vectors. We describe an algorithm for updating the \csr as an agent moves and connect the representation to the agent itself.
We discuss use of our pipeline to tackle visual room rearrangement.
Our experiments show
(1) A simple planner using \csr outperforms sophisticated baselines on visual room rearrangement, even though our model is not trained directly for this task.
(2) The relational feature from \csr can be used to recover semantically meaningful spatial relations.
(3) The representation can detect subtle visual changes in the scene.
We hope this work will encourage investigation of techniques in scene representation learning for other downstream embodied tasks, going beyond learning task specific representations.

{\small \noindent \textbf{Acknowledgements.} We would like to thank Winson Han for help with figures, Klemen Kotar for detector weights, and Luca Weihs for help with AllenAct.
SS is supported by NSF CMMI-2037101 and NSF IIS-2132519. The views and conclusions contained herein are those of the authors and should not be interpreted as necessarily representing the official policies, either expressed or implied, of the sponsors.
}
{\small
\bibliographystyle{ieee_fullname}
\bibliography{egbib}

\begin{thebibliography}{10}\itemsep=-1pt

\bibitem{anderson2018vision}
Peter Anderson, Qi Wu, Damien Teney, Jake Bruce, Mark Johnson, Niko
  S{\"u}nderhauf, Ian Reid, Stephen Gould, and Anton Van Den~Hengel.
\newblock Vision-and-language navigation: Interpreting visually-grounded
  navigation instructions in real environments.
\newblock In {\em CVPR}, 2018.

\bibitem{armeni20193d}
Iro Armeni, Zhi-Yang He, JunYoung Gwak, Amir~R Zamir, Martin Fischer, Jitendra
  Malik, and Silvio Savarese.
\newblock 3d scene graph: A structure for unified semantics, 3d space, and
  camera.
\newblock In {\em ICCV}, 2019.

\bibitem{bao2012semantic}
Sid~Yingze Bao, Mohit Bagra, Yu-Wei Chao, and Silvio Savarese.
\newblock Semantic structure from motion with points, regions, and objects.
\newblock In {\em CVPR}, 2012.

\bibitem{batra2020rearrangement}
Dhruv Batra, Angel~X Chang, Sonia Chernova, Andrew~J Davison, Jia Deng, Vladlen
  Koltun, Sergey Levine, Jitendra Malik, Igor Mordatch, Roozbeh Mottaghi,
  et~al.
\newblock Rearrangement: A challenge for embodied ai.
\newblock {\em arXiv}, 2020.

\bibitem{objectnav}
Dhruv Batra, Aaron Gokaslan, Aniruddha Kembhavi, Oleksandr Maksymets, Roozbeh
  Mottaghi, Manolis Savva, Alexander Toshev, and Erik Wijmans.
\newblock Objectnav revisited: On evaluation of embodied agents navigating to
  objects.
\newblock {\em arXiv}, 2020.

\bibitem{Battaglia2016InteractionNF}
Peter~W. Battaglia, Razvan Pascanu, Matthew Lai, Danilo~Jimenez Rezende, and
  Koray Kavukcuoglu.
\newblock Interaction networks for learning about objects, relations and
  physics.
\newblock In {\em NeurIPS}, 2016.

\bibitem{bowman2017probabilistic}
Sean~L Bowman, Nikolay Atanasov, Kostas Daniilidis, and George~J Pappas.
\newblock Probabilistic data association for semantic slam.
\newblock In {\em ICRA}, 2017.

\bibitem{chaplot2020learning}
Devendra~Singh Chaplot, Dhiraj Gandhi, Saurabh Gupta, Abhinav Gupta, and Ruslan
  Salakhutdinov.
\newblock Learning to explore using active neural slam.
\newblock In {\em ICLR}, 2020.

\bibitem{chaplot2020neural}
Devendra~Singh Chaplot, Ruslan Salakhutdinov, Abhinav Gupta, and Saurabh Gupta.
\newblock Neural topological slam for visual navigation.
\newblock In {\em CVPR}, 2020.

\bibitem{Savarese-RSS-19}
Kevin Chen, Juan~Pablo de Vicente, Gabriel Sepulveda, Fei Xia, Alvaro Soto,
  Marynel Vazquez, and Silvio Savarese.
\newblock A behavioral approach to visual navigation with graph localization
  networks.
\newblock In {\em RSS}, 2019.

\bibitem{chen2020improved}
Xinlei Chen, Haoqi Fan, Ross Girshick, and Kaiming He.
\newblock Improved baselines with momentum contrastive learning.
\newblock {\em arXiv}, 2020.

\bibitem{chen2019holistic++}
Yixin Chen, Siyuan Huang, Tao Yuan, Siyuan Qi, Yixin Zhu, and Song-Chun Zhu.
\newblock Holistic++ scene understanding: Single-view 3d holistic scene parsing
  and human pose estimation with human-object interaction and physical
  commonsense.
\newblock In {\em ICCV}, 2019.

\bibitem{cong2021spatial}
Yuren Cong, Wentong Liao, Hanno Ackermann, Bodo Rosenhahn, and Michael~Ying
  Yang.
\newblock Spatial-temporal transformer for dynamic scene graph generation.
\newblock In {\em ICCV}, 2021.

\bibitem{dai2017detecting}
Bo Dai, Yuqi Zhang, and Dahua Lin.
\newblock Detecting visual relationships with deep relational networks.
\newblock In {\em CVPR}, 2017.

\bibitem{Das2018EmbodiedQA}
Abhishek Das, Samyak Datta, Georgia Gkioxari, Stefan Lee, Devi Parikh, and
  Dhruv Batra.
\newblock Embodied question answering.
\newblock In {\em CVPR}, 2018.

\bibitem{davison2007monoslam}
Andrew~J Davison, Ian~D Reid, Nicholas~D Molton, and Olivier Stasse.
\newblock Monoslam: Real-time single camera slam.
\newblock {\em TPAMI}, 2007.

\bibitem{du2020learning}
Heming Du, Xin Yu, and Liang Zheng.
\newblock Learning object relation graph and tentative policy for visual
  navigation.
\newblock In {\em ECCV}, 2020.

\bibitem{ehsani2021manipulathor}
Kiana Ehsani, Winson Han, Alvaro Herrasti, Eli VanderBilt, Luca Weihs, Eric
  Kolve, Aniruddha Kembhavi, and Roozbeh Mottaghi.
\newblock Manipulathor: A framework for visual object manipulation.
\newblock In {\em CVPR}, 2021.

\bibitem{engel2014lsd}
Jakob Engel, Thomas Sch{\"o}ps, and Daniel Cremers.
\newblock Lsd-slam: Large-scale direct monocular slam.
\newblock In {\em ECCV}, 2014.

\bibitem{fisher2011characterizing}
Matthew Fisher, Manolis Savva, and Pat Hanrahan.
\newblock Characterizing structural relationships in scenes using graph
  kernels.
\newblock In {\em ACM SIGGRAPH}, 2011.

\bibitem{gadre2021act}
Samir~Yitzhak Gadre, Kiana Ehsani, and Shuran Song.
\newblock Act the part: Learning interaction strategies for articulated object
  part discovery.
\newblock {\em ICCV}, 2021.

\bibitem{gan2021threedworld}
Chuang Gan, Jeremy Schwartz, Seth Alter, Damian Mrowca, Martin Schrimpf, James
  Traer, Julian~De Freitas, Jonas Kubilius, Abhishek Bhandwaldar, Nick Haber,
  Megumi Sano, Kuno Kim, Elias Wang, Michael Lingelbach, Aidan Curtis,
  Kevin~Tyler Feigelis, Daniel Bear, Dan Gutfreund, David~Daniel Cox, Antonio
  Torralba, James~J. DiCarlo, Joshua~B. Tenenbaum, Josh Mcdermott, and
  Daniel~LK Yamins.
\newblock Three{DW}orld: A platform for interactive multi-modal physical
  simulation.
\newblock In {\em NeurIPS}, 2021.

\bibitem{Gordon2018IQAVQ}
Daniel Gordon, Aniruddha Kembhavi, Mohammad Rastegari, Joseph Redmon, Dieter
  Fox, and Ali Farhadi.
\newblock Iqa: Visual question answering in interactive environments.
\newblock In {\em CVPR}, 2018.

\bibitem{Gupta2019LVISAD}
Agrim Gupta, Piotr Doll{\'a}r, and Ross~B. Girshick.
\newblock Lvis: A dataset for large vocabulary instance segmentation.
\newblock {\em CVPR}, 2019.

\bibitem{gupta2017cognitive}
Saurabh Gupta, James Davidson, Sergey Levine, Rahul Sukthankar, and Jitendra
  Malik.
\newblock Cognitive mapping and planning for visual navigation.
\newblock In {\em CVPR}, 2017.

\bibitem{he2020momentum}
Kaiming He, Haoqi Fan, Yuxin Wu, Saining Xie, and Ross Girshick.
\newblock Momentum contrast for unsupervised visual representation learning.
\newblock {\em CVPR}, 2020.

\bibitem{hubert1985comparing}
L. Hubert and P. Arabie.
\newblock Comparing partitions.
\newblock {\em Journal of Classification}, 1985.

\bibitem{ji2020action}
Jingwei Ji, Ranjay Krishna, Li Fei-Fei, and Juan~Carlos Niebles.
\newblock Action genome: Actions as compositions of spatio-temporal scene
  graphs.
\newblock In {\em CVPR}, 2020.

\bibitem{Johnson2015CVPR}
Justin Johnson, Ranjay Krishna, Michael Stark, Li-Jia Li, David~A Shamma,
  Michael Bernstein, and Li Fei-Fei.
\newblock Image retrieval using scene graphs.
\newblock In {\em CVPR}, 2015.

\bibitem{ai2thor}
Eric Kolve, Roozbeh Mottaghi, Winson Han, Eli VanderBilt, Luca Weihs, Alvaro
  Herrasti, Daniel Gordon, Yuke Zhu, Abhinav Gupta, and Ali Farhadi.
\newblock {AI2-THOR: An Interactive 3D Environment for Visual AI}.
\newblock {\em arXiv}, 2017.

\bibitem{li2021embodied}
Xinghang Li, Di Guo, Huaping Liu, and Fuchun Sun.
\newblock Embodied semantic scene graph generation.
\newblock In {\em CoRL}, 2021.

\bibitem{li2017scene}
Yikang Li, Wanli Ouyang, Bolei Zhou, Kun Wang, and Xiaogang Wang.
\newblock Scene graph generation from objects, phrases and region captions.
\newblock In {\em ICCV}, 2017.

\bibitem{liu2020beyond}
Chenchen Liu, Yang Jin, Kehan Xu, Guoqiang Gong, and Yadong Mu.
\newblock Beyond short-term snippet: Video relation detection with
  spatio-temporal global context.
\newblock In {\em CVPR}, 2020.

\bibitem{lu2016visual}
Cewu Lu, Ranjay Krishna, Michael Bernstein, and Li Fei-Fei.
\newblock Visual relationship detection with language priors.
\newblock In {\em ECCV}, 2016.

\bibitem{mur2017orb}
Raul Mur-Artal and Juan~D Tard{\'o}s.
\newblock Orb-slam2: An open-source slam system for monocular, stereo, and
  rgb-d cameras.
\newblock {\em IEEE Trans. on Robotics}, 2017.

\bibitem{ost2021neural}
Julian Ost, Fahim Mannan, Nils Thuerey, Julian Knodt, and Felix Heide.
\newblock Neural scene graphs for dynamic scenes.
\newblock In {\em CVPR}, 2021.

\bibitem{perez2021robot}
Claudia P{\'e}rez-D’Arpino, Can Liu, Patrick Goebel, Roberto
  Mart{\'\i}n-Mart{\'\i}n, and Silvio Savarese.
\newblock Robot navigation in constrained pedestrian environments using
  reinforcement learning.
\newblock In {\em ICRA}, 2021.

\bibitem{ren2015faster}
Shaoqing Ren, Kaiming He, Ross Girshick, and Jian Sun.
\newblock Faster r-cnn: Towards real-time object detection with region proposal
  networks.
\newblock In {\em NeurIPS}, 2015.

\bibitem{rosinol20203d}
Antoni Rosinol, Arjun Gupta, Marcus Abate, Jingnan Shi, and Luca Carlone.
\newblock 3d dynamic scene graphs: Actionable spatial perception with places,
  objects, and humans.
\newblock In {\em RSS}, 2020.

\bibitem{Rosinol21ijrr-Kimera}
A. Rosinol, A. Violette, M. Abate, N. Hughes, Y. Chang, J. Shi, A. Gupta, and
  L. Carlone.
\newblock Kimera: from {SLAM} to spatial perception with {3D} dynamic scene
  graphs.
\newblock {\em Intl. J. of Robotics Research}, 2021.

\bibitem{runz2017co}
Martin R{\"u}nz and Lourdes Agapito.
\newblock Co-fusion: Real-time segmentation, tracking and fusion of multiple
  objects.
\newblock In {\em ICRA}, 2017.

\bibitem{salas2013slam++}
Renato~F Salas-Moreno, Richard~A Newcombe, Hauke Strasdat, Paul~HJ Kelly, and
  Andrew~J Davison.
\newblock Slam++: Simultaneous localisation and mapping at the level of
  objects.
\newblock In {\em CVPR}, 2013.

\bibitem{savinov2018semi}
Nikolay Savinov, Alexey Dosovitskiy, and Vladlen Koltun.
\newblock Semi-parametric topological memory for navigation.
\newblock In {\em ICLR}, 2018.

\bibitem{Schulman2017ProximalPO}
John Schulman, Filip Wolski, Prafulla Dhariwal, Alec Radford, and Oleg Klimov.
\newblock Proximal policy optimization algorithms.
\newblock {\em arXiv}, 2017.

\bibitem{shang2017video}
Xindi Shang, Tongwei Ren, Jingfan Guo, Hanwang Zhang, and Tat-Seng Chua.
\newblock Video visual relation detection.
\newblock In {\em ACM Multimedia}, 2017.

\bibitem{shen2020igibson}
Bokui Shen, Fei Xia, Chengshu Li, Roberto Mart{\'\i}n-Mart{\'\i}n, Linxi Fan,
  Guanzhi Wang, Shyamal Buch, Claudia D'Arpino, Sanjana Srivastava, Lyne~P
  Tchapmi, et~al.
\newblock igibson, a simulation environment for interactive tasks in large
  realistic scenes.
\newblock In {\em IROS}, 2021.

\bibitem{Shridhar2020ALFREDAB}
Mohit Shridhar, Jesse Thomason, Daniel Gordon, Yonatan Bisk, Winson Han,
  Roozbeh Mottaghi, Luke Zettlemoyer, and Dieter Fox.
\newblock Alfred: A benchmark for interpreting grounded instructions for
  everyday tasks.
\newblock In {\em CVPR}, 2020.

\bibitem{strecke2019fusion}
Michael Strecke and Jorg Stuckler.
\newblock Em-fusion: Dynamic object-level slam with probabilistic data
  association.
\newblock In {\em ICCV}, 2019.

\bibitem{tsai2019video}
Yao-Hung~Hubert Tsai, Santosh Divvala, Louis-Philippe Morency, Ruslan
  Salakhutdinov, and Ali Farhadi.
\newblock Video relationship reasoning using gated spatio-temporal energy
  graph.
\newblock In {\em CVPR}, 2019.

\bibitem{oord2018representation}
A{\"a}ron van~den Oord, Yazhe Li, and Oriol Vinyals.
\newblock Representation learning with contrastive predictive coding.
\newblock {\em arXiv}, 2018.

\bibitem{wald2020learning}
Johanna Wald, Helisa Dhamo, Nassir Navab, and Federico Tombari.
\newblock Learning 3d semantic scene graphs from 3d indoor reconstructions.
\newblock In {\em CVPR}, 2020.

\bibitem{weihs2021visual}
Luca Weihs, Matt Deitke, Aniruddha Kembhavi, and Roozbeh Mottaghi.
\newblock Visual room rearrangement.
\newblock In {\em CVPR}, 2021.

\bibitem{Weihs2020AllenActAF}
Luca Weihs, Jordi Salvador, Klemen Kotar, Unnat Jain, Kuo-Hao Zeng, Roozbeh
  Mottaghi, and Aniruddha Kembhavi.
\newblock Allenact: A framework for embodied ai research.
\newblock {\em arXiv}, 2020.

\bibitem{Wijmans2020DDPPOLN}
Erik Wijmans, Abhishek Kadian, Ari~S. Morcos, Stefan Lee, Irfan Essa, Devi
  Parikh, Manolis Savva, and Dhruv Batra.
\newblock Dd-ppo: Learning near-perfect pointgoal navigators from 2.5 billion
  frames.
\newblock In {\em ICLR}, 2020.

\bibitem{wong2021rigidfusion}
Yu-Shiang Wong, Changjian Li, Matthias Niessner, and Niloy~J. Mitra.
\newblock Rigidfusion: Rgb-d scene reconstruction with rigidly-moving objects.
\newblock {\em Computer Graphics Forum}, 2021.

\bibitem{wortsman2019learning}
Mitchell Wortsman, Kiana Ehsani, Mohammad Rastegari, Ali Farhadi, and Roozbeh
  Mottaghi.
\newblock Learning to learn how to learn: Self-adaptive visual navigation using
  meta-learning.
\newblock In {\em CVPR}, 2019.

\bibitem{wu2019detectron2}
Yuxin Wu, Alexander Kirillov, Francisco Massa, Wan-Yen Lo, and Ross Girshick.
\newblock Detectron2.
\newblock \url{https://github.com/facebookresearch/detectron2}, 2019.

\bibitem{wu2019bayesian}
Yi Wu, Yuxin Wu, Aviv Tamar, Stuart Russell, Georgia Gkioxari, and Yuandong
  Tian.
\newblock Bayesian relational memory for semantic visual navigation.
\newblock In {\em ICCV}, 2019.

\bibitem{xiang2018posecnn}
Yu Xiang, Tanner Schmidt, Venkatraman Narayanan, and Dieter Fox.
\newblock Posecnn: A convolutional neural network for 6d object pose estimation
  in cluttered scenes.
\newblock {\em RSS}, 2018.

\bibitem{xu2019mid}
Binbin Xu, Wenbin Li, Dimos Tzoumanikas, Michael Bloesch, Andrew Davison, and
  Stefan Leutenegger.
\newblock Mid-fusion: Octree-based object-level multi-instance dynamic slam.
\newblock In {\em ICRA}, 2019.

\bibitem{xu2017scene}
Danfei Xu, Yuke Zhu, Christopher~B Choy, and Li Fei-Fei.
\newblock Scene graph generation by iterative message passing.
\newblock In {\em CVPR}, 2017.

\bibitem{yang2018graph}
Jianwei Yang, Jiasen Lu, Stefan Lee, Dhruv Batra, and Devi Parikh.
\newblock Graph r-cnn for scene graph generation.
\newblock In {\em ECCV}, 2018.

\bibitem{yang2018visual}
Wei Yang, Xiaolong Wang, Ali Farhadi, Abhinav Gupta, and Roozbeh Mottaghi.
\newblock Visual semantic navigation using scene priors.
\newblock In {\em ICLR}, 2019.

\bibitem{zellers2018neural}
Rowan Zellers, Mark Yatskar, Sam Thomson, and Yejin Choi.
\newblock Neural motifs: Scene graph parsing with global context.
\newblock In {\em CVPR}, 2018.

\bibitem{zhao2013scene}
Yibiao Zhao and Song-Chun Zhu.
\newblock Scene parsing by integrating function, geometry and appearance
  models.
\newblock In {\em CVPR}, 2013.

\bibitem{zhou2019scenegraphnet}
Yang Zhou, Zachary While, and Evangelos Kalogerakis.
\newblock Scenegraphnet: Neural message passing for 3d indoor scene
  augmentation.
\newblock In {\em ICCV}, 2019.

\bibitem{zhu2021hierarchical}
Yifeng Zhu, Jonathan Tremblay, Stan Birchfield, and Yuke Zhu.
\newblock Hierarchical planning for long-horizon manipulation with geometric
  and symbolic scene graphs.
\newblock In {\em ICRA}, 2021.

\end{thebibliography}
}

\clearpage
\appendix
\section*{Appendix}
\section{Faster-RCNN}
\label{appx:rcnn}
Here we mention the architecture, data, and training details for the Faster-RCNN network.

\mypara{Architecture.}
We use the Faster-RCNN default architecture from Detectron2 \cite{wu2019detectron2}.

\mypara{Data.}
We use $\sim$10k images from train rooms in AI2-THOR with bounding box labels and $\sim$127k train images from LVIS \cite{Gupta2019LVISAD}.
We choose this number of THOR images to ensure diversity of views without repetition of frames that would appear very similar.
In total the detector is trained on 1235 classes (accounting for the class overlap between LVIS and the $\sim$100 THOR categories).

\mypara{Training.}
We train using the default Detectron2 3x schedule.
We train on a machine with eight GeForce GTX TITAN X NVIDIA GPUs.
The training takes $\sim$2 days.

\mypara{Inference.}
At inference, we treat the Faster-RCNN module as a region proposal network. 
Hence, we only require its detections for extracting node and edge features.
Our proposed algorithm does not use the predicted class labels as input to create node and edge feature representations.

\section{Policy}
\label{appx:policy}
Here we mention the architecture, data, and training details for the policy network.

\mypara{Architecture.}
The policy consists of three convolutional layers and a GRU.
It is an actor-critic style network.
The input is an image $I \in \mathbb{R}^{224 \times 224 \times 3}$.
Note, we do not feed in the action from the last timestep.
The output of the conv. backbone is a volume, $v \in \mathbb{R}^{24 \times 24 \times 512}$, which is flattened and projected by a linear layer into a feature $x \in \mathbb{R}^{512}$.
$x$ is taken as input to the GRU module, which maintains a hidden state $h \in \mathbb{R}^{512}$ and outputs another feature vector $z \in \mathbb{R}^{512}$.
An linear actor head takes $z$ and projects it to give logits over the eight discrete actions.
A second linear critic head takes $z$ and projects it to give the critic score.

\mypara{Data.}
We conduct training rollouts within random starting locations drawn from the 80 train rooms in AI2-THOR.

\mypara{Training.}
We adopt the AllenAct \cite{Weihs2020AllenActAF} framework for training.
Specifically we use the DD-PPO \cite{Schulman2017ProximalPO, Wijmans2020DDPPOLN} algorithm to train our network.
We train on machines that have 48 CPU cores and four T4 NVIDIA GPUs.
We train for 200 million steps, which takes  $\sim$2 days.
We use default AllenAct PPO settings, with rollout episode length of 150 steps.
We employ sparse rewards, which are computed based on known simulation state at training.
The agent receives positive reward of 0.1 if it visits a new position (agent orientation is disregarded) and reward of 0.4 if it sees a new object within a rollout.
There is also a step penalty of -0.01 and a failed action penalty of -0.03.

\section{Continuous Scene Representation}
\label{appx:csr}

\mypara{Architecture.}
We use a standard ResNet-18 architecture.
We modify \texttt{conv1} to take five channel input (three channels for RGB, a forth channel for the first binary box mask, and a fifth channel for the second binary box mask).
Hence our ResNet takes input $\in \mathbb{R}^{224 \times 224 \times 5}$.
After the ResNet, we have an MLP bottleneck projection head, which takes in a feature $\in \mathbb{R}^{512}$ and outputs a feature $\in \mathbb{R}^{512}$.
Architectural details for the network that extracts object correspondence features are the same.

\mypara{Data.}
As stated in the paper we capture 20 random agent poses in 5 different configurations (random object placement and scene textures) in the 80 different train rooms. This leads to a train dataset of $\sim$600k relations.
The rest of the dataset is composed of a near even split between validation and test relations, yielding a total dataset size of $\sim$900k relations.
We provide some train statistics to give a better understanding of the dataset.
Of the $\sim$600k relations, $\sim$60k are node relations, the rest are directed edge relations.
In total there are $\sim$3k different object instances across the $\sim$100 AI2-THOR categories.

\mypara{Training.}
We train on a machine with eight GeForce GTX TITAN X NVIDIA GPUs.
Our learning leverages InfoNCE \cite{oord2018representation} loss and builds on the MoCo framework \cite{he2020momentum, chen2020improved}.
We use a relatively small queue of size of 1024 for negatives.
The InfoNCE temperature parameter is 0.07 and the momentum update coefficient is 0.999.
We train in minibatches of 512, with initial learning rate of 0.1, a cosine decay schedule, and standard SGD w/ momentum optimizer.
Our model take less than one day to converge.

\mypara{Constants.}
We must set three thresholds in our method (1) to determine within trajectory matches, (2) to determine object correspondences matches, and (3) to determine if an object has moved.
If (1) cosine similarity between two matched node features is greater than 0.5 within a trajectory, we consider the instances as a true match.
If cosine similarity of matched object features between trajectories is greater than 0.4, we consider the objects to be a true match.
Finally, if cosine similarity of the node features drops below 0.8 after nodes have been matched via object features between trajectories, we consider the object to be a candidate object that has moved.

\section{Linear Probes}
Here we provide more relevant details for our linear probes.

\mypara{Data.}
We consider two tasks, \textsc{[support]} and \textsc{[sibling]}. For \textsc{[support]} we create a balanced dataset with $\sim$2k positive examples of an object on top of another object.
By reversing the order of the boxes for the input, we get another $\sim$2k examples of an object under another object.
Finally we create a third category of $\sim$2k examples of unrelated objects (i.e., objects that do not follow the \textsc{[support]} relationship).
For the \textsc{[sibling]} relation we create two categories each with $\sim$2k examples, the first with examples from this relationship (i.e., two objects on the same receptacle) and the second of unrelated objects (i.e., objects that do not follow the \textsc{[sibling]} relationship).
For both datasets we use a 80/20 train/test split for each category.

\mypara{Training.}
For training our model, we conduct a linear probe, with the high learning rate of $0.5$, using the validation set loss to determine convergence.
We use features before the MLP projection head as is common for linear probes in the contrastive learning literature.
For end-to-end baseline, we first train with reasonable parameters (i.e., learning rate of 0.02, with cosine schedule, SGD w/ momentum, weight decay of 0.001) for 100 epochs, taking the checkpoint with the lowest validation loss.
We then use the same linear probe routine discussed above to probe the baseline representations for transfer performance.
\begin{figure}[t]
    \centering
    \includegraphics[width=\linewidth]{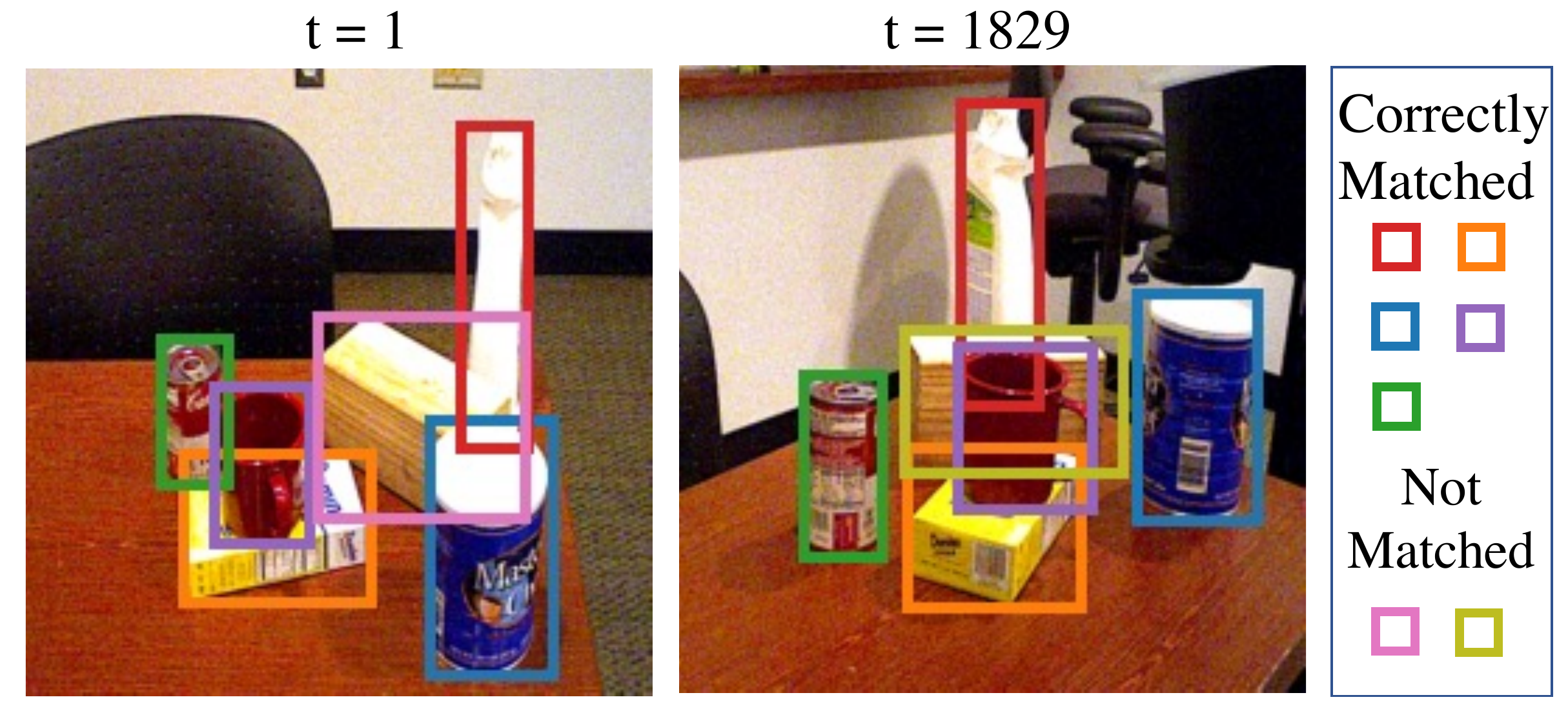}
    % \vspace{-2em}
    \caption{
        \textbf{Qualitative matching on YCB-Video.}. All but a heavily occluded.
    }
    \label{fig:real_world_qual}
\end{figure}

\section{Exploration Heuristic}

To ablate the effectiveness of our exploration policy in the visual room rearrangement pipeline, we also design a heuristic policy.
Given the simulation state of the room before and after the shuffle, we can retrieve the squares in the map that are closest (in terms of euclidean distance) to the objects that get shuffled.
Hence, we get $2n$ locations, where $n$ is the number of objects that get shuffled.
During the walkthrough trajectory, based on the agent's current location, the heuristic policy greedily picks the closest location and takes the shortest path to this point.
During the unshuffle exploration trajectory the locations are visited in reversed order (e.g., the waypoint visited last in the walkthrough is visited first in the unshuffle).

\section{AI2-THOR Assets}
AI2-THOR assets are available under Apache 2.0.

\section{Rearrangement}
While our setting is identical to that of Weihs \textit{et al.} \cite{weihs2021visual}, our method does not attempt to fix objects that have changed state (e.g., drawers opening).
Hence our method cannot successfully rearrange rooms with these changes.
However, for fair comparison to prior work, we report numbers on the full RoomR dataset with all data points.
Planning for objects that change in openness is left to future work.

\section{Qualitative Real World Tracking Results}

We show a qualitative example where all but one heavily occluded object is properly matched in Fig.~\ref{fig:real_world_qual}.

\end{document}